\documentclass[10pt,twocolumn,letterpaper]{article}

\usepackage[pagenumbers]{cvpr} 

\usepackage{graphicx}
\usepackage{amsmath}
\usepackage{amssymb}
\usepackage{booktabs}

\usepackage{textcomp}
\usepackage{gensymb}
\usepackage[pagebackref,breaklinks,colorlinks]{hyperref}
\usepackage{multirow}

\usepackage[capitalize]{cleveref}
\crefname{section}{Sec.}{Secs.}
\Crefname{section}{Section}{Sections}
\Crefname{table}{Table}{Tables}
\crefname{table}{Tab.}{Tabs.}

\begin{document}

\title{Pixel-wise Deep Image Stitching}

\author{
\textbf{Hyeokjun Kweon\thanks{The first three authors contributed equally. In alphabetical order.}}
\quad
\textbf{Hyeonseong Kim$^{*}$}
\quad 
\textbf{Yoonsu Kang$^{*}$}
\\
\textbf{Youngho Yoon}
\quad
\textbf{Wooseong Jeong}
\quad
\textbf{Kuk-Jin Yoon}
\and
Visual Intelligence Lab., KAIST, Korea\\
{\tt\small \{0327june, brian617, gzgzys9887, dudgh1732, stk14570, kjyoon\}@kaist.ac.kr}
}
\maketitle

\begin{abstract}
   Image stitching aims at stitching the images taken from different viewpoints into an image with a wider field of view.
   Existing methods warp the target image to the reference image using the estimated warp function, and a homography is one of the most commonly used warping functions. 
   However, when images have large parallax due to non-planar scenes and translational motion of a camera, the homography cannot fully describe the mapping between two images.
   Existing approaches based on global or local homography estimation are not free from this problem and suffer from undesired artifacts due to parallax.
   In this paper, instead of relying on the homography-based warp, we propose a novel deep image stitching framework exploiting the pixel-wise warp field to handle the large-parallax problem.
   The proposed deep image stitching framework consists of two modules: Pixel-wise Warping Module (PWM) and Stitched Image Generating Module (SIGMo).
   PWM employs an optical flow estimation model to obtain pixel-wise warp of the whole image, and relocates the pixels of the target image with the obtained warp field.
   SIGMo blends the warped target image and the reference image while eliminating unwanted artifacts such as misalignments, seams, and holes that harm the plausibility of the stitched result.
   For training and evaluating the proposed framework, we build a large-scale dataset that includes image pairs with corresponding pixel-wise ground truth warp and sample stitched result images.
   We show that the results of the proposed framework are qualitatively superior to those of the conventional methods, especially when the images have large parallax. The code and the proposed dataset will be publicly available soon.
   
\end{abstract}

\begin{figure}[ht]
    \centering
    \includegraphics[width=0.99\linewidth]{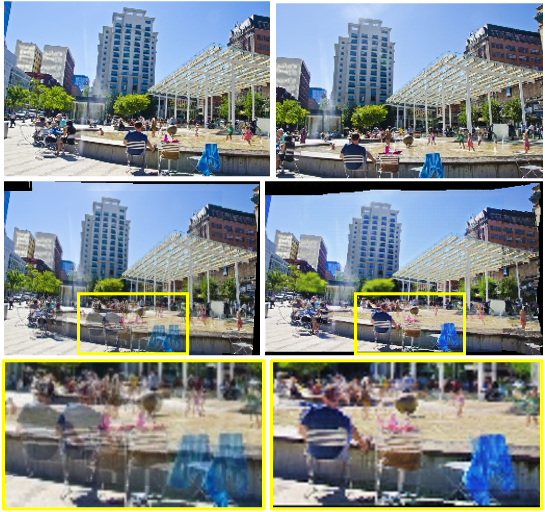}
    \vspace{-7pt}
    \caption{Qualitative comparison on large parallax images. Reference image (left top), target image (right top), stitched image of APAP~\cite{apap} (left bottom), and stitched image of our method (right bottom). More qualitative results are in Fig.~\ref{fig:lapis}.}
    \label{fig:intro}
    \vspace{-10pt}
\end{figure}

\section{Introduction}
Image stitching is a classic computer vision task widely used in diverse applications such as robot navigation or panorama/${360^\circ}$ image acquisition.
Image stitching aligns multiple images taken from different viewpoints into an image from a specific viewpoint in order to obtain an image with a wider field of view.
A general pipeline of image stitching methods can be described  as a series of sub-tasks: 1) searching for correspondences between two images, 2) obtaining transformation between two images based on the correspondences, usually a homography-based warping function, 3) warping an image with the transformation, and 4) blending the images to reduce unpleasant seams~\cite{gradient,seagull,largemotion} or projective distortions~\cite{sphp,aanap,li2017quasi}.

AutoStitch~\cite{autostitch}, one of the earliest works, used keypoint detection and matching to obtain point correspondences between two images.
Based on the correspondences, homography transformation between two images is optimized under the planar scene assumption.
However, this assumption cannot be established when the camera and the target scene are close and/or scenes include abrupt depth changes.
In these cases, misalignment named as ``parallax problem" occurs and leads to unpleasant effects such as ghosting artifacts in the stitched images.

To address this problem, several approaches~\cite{DHW, apap, PCPS} have proposed to divide an input image into subregions and then optimize a transformation matrix for each cell while assuming that each subregion is planar.
However, there is a trade-off between accuracy and convergence of warping function when using these approaches.
If we divide the image into smaller subregions to make them nearly planar, then the correspondences on such subregions would be scarce and therefore difficult to optimize the regional warp on these regions.
On the other hand, if we use larger subregions, it would provide sufficient matchings but the planar assumption could be weakened.
\cite{PCPS} attempts to segment the target image into triangles and merges them into projective-consistent planes. 
However, it still assumes continuous scene depths and relies on the estimated depth.

Our solution for this dilemma is to break away from the paradigm of regional warping using homography.
Instead of estimating the warp in the form of a regionally shared transformation matrix, we propose to directly estimate the 2D pixel-wise warp field for image stitching, similar to optical flow.
Since this approach defines neither subregions nor regional warp, it is free from the aforementioned trade-off and therefore can handle large-parallax scenes.
In this paper, we propose a novel pixel-wise deep image stitching framework and dataset.
Specifically, we design two modules: Pixel-wise Warping Module (PWM) and Stitched Image Generating Module (SIGMo).

PWM aims at estimating and applying pixel-wise warp from the target image to the reference image.
In the PWM, a warp estimation network is trained to estimate the pixel-wise warp, with the help of the ground truth (GT) supervision.
In the overlap (OV) region, as similar to optical flow estimation, PWM can learn to estimate pixel-wise warp based on the point correspondences.
On the other hand, acquiring pixel-wise warp for the non-overlap (NOV) region is an ill-defined problem due to the absence of correspondences.
From the perspective of image stitching, warp leading to a plausible stitching result is not unique.
For this reason, instead of imposing an identical loss function on both the OV and NOV regions, we regularize the loss for NOV regions.
It helps the model learn a tendency of pixel-wise warp over the NOV region during the training while preventing the model from over-fitting on the training dataset.
With the acquired pixel-wise warp, we relocate the pixels of the target image with forward warping.

SIGMo focuses on generating a plausible stitching result by blending the warped target image from PWM and the reference image.
Since the warped target image includes holes caused by forward warping, we make SIGMo fill the holes while handling the other undesired artifacts such as seams or misalignments.
To achieve this, we devise hole-aware reconstruction loss which preserves the reliable pixels of the reference and warped target images.
In addition, to reduce the undesired artifacts from the stitched result, we exploit adversarial loss from the unconditional discriminator.

One of the biggest obstacles to introducing deep learning to the field of image stitching is the absence of the dataset.
A few existing studies on image stitching~\cite{apap, ptis} have published datasets composed of multi-view images for qualitative evaluation purpose. 
However, these sets are insufficient to train a deep learning model since they contain only tens of images and also do not include stitched images due to the nature of the stitching task.
Recently, a large-scale dataset~\cite{uis} for image stitching was proposed.
However, the set also contains neither GT point correspondences nor stitched images since the study targets unsupervised setting.
Therefore, to train and evaluate the proposed framework, we build a large-scale synthetic dataset based on a photo-realistic virtual environment~\cite{stanford2d3d-dataset}.
This dataset includes pairs of reference and target images, where the ground truth pixel-wise warp fields and sample stitched images are also provided.

Our approach achieves substantial results that outperform the existing image stitching methods on both the proposed dataset and the real images.
As shown in Fig.~\ref{fig:intro}, we observed that the proposed method effectively performs image stitching, especially for the scenes containing large parallax or non-planar objects where the existing approaches usually have failed.
%
In summary, our contributions are as follows:

\noindent (1) We propose to estimate pixel-wise warp to address the large-parallax problem in the image stitching task.

\noindent (2) We develop a novel deep image stitching framework that estimates pixel-wise warp and blends the warped target image with a reference image.

\noindent (3) We propose a large-scale dataset for the training and evaluation of the image stitching task.

\noindent (4) We obtain superior qualitative stitching results on both the proposed dataset and real images with the proposed framework compared to the existing methods, especially for the scenes including large-parallax.

\section{Related Work}

\noindent\textbf{Image Stitching}
Existing image stitching approaches have proposed to divide an input image into subregions and then optimize a transformation matrix for each cell while assuming that the subregions are planar. 

\cite{DHW} divided the scene into distant plane and ground plane then optimized dual homography for each plane. 
\cite{apap} proposed to divide the target image into cell-like subregions and optimize a homography matrix for each cell with the help of moving direct linear transformation (MDLT) algorithm.
\cite{PCPS} attempted to segment the target image into triangles and merge them into the projective-consistent planes.
While minimizing the warping residual for each feature,~\cite{lee2020warping} divided input images into superpixels and then adaptively warp each superpixel.
However, using a regional warp for each subregion induces a trade-off between the accuracy and convergence of the optimization for warp.
Also, the additional computational cost required for partitioning the image and optimizing homography for each region cannot be negligible.
Some methods have attempted to refine the homography-based warp with various constraints.
\cite{sva} refined an initial global warp with a spatially-varying affine transformation.
\cite{zhang2016multi} devised an energy function to minimize the distortion of local scale and lines in the stitching result.
\cite{lee2020warping} proposed an analytical warping function and warp the input image over the meshed image plane to eliminate the parallax errors caused by the homography.
\cite{jia2021leveraging} proposed structure-preserving image stitching method with a warping guided by line-point constraint.

These works, however, cannot fully resolve the drawback of using homography-based warp since the refined warp is a locally adjusted version of the initial warp.
The proposed framework, on the other hand, estimates the pixel-wise warp instead of using homography.

\vspace{3pt}
\noindent\textbf{Optical Flow} 
Due to the recent advances in deep learning, existing deep optical flow (OF) networks have produced successful results. 
End-to-end optical flow estimation approaches search for point correspondences, and therefore only target flow estimation in the overlap region~\cite{flownet2, pwc, raft, gma}.
Yet, non-overlap flow estimation is an ill-posed problem as there are no point correspondences in the non-overlap region.
However, acquiring the transformation function of non-overlap to overlap region domain is necessary for image stitching.
Since we employ a pixel-wise warp-based method instead of homography, we need to address flow estimation in the non-overlap region.

\begin{figure*}[ht]
    \centering
    \includegraphics[width=0.99\linewidth]{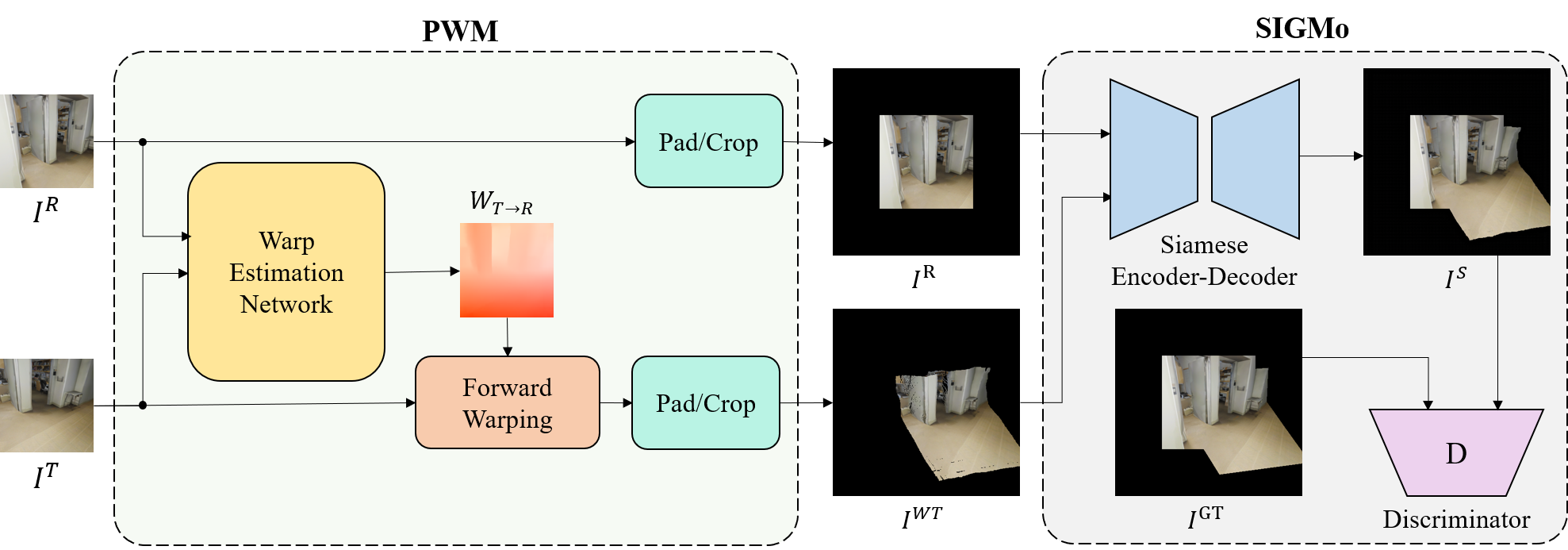}
    \vspace{-7pt}
    \caption{Overall framework for pixel-wise deep image stitching. Given $I^R$ and $I^T$, PWM relocates the pixels in $I^T$ to the $I^R$ domain using the estimated warp field $W_{T \rightarrow R}$ in order to obtain $I^{WT}$. Then, $I^R$ and $I^{WT}$ are padded or cropped to match the predefined size of the stitched image. Finally, $I^R$ and $I^{WT}$ are fed into the SIGMo to obtain a stitched image $I^S$.
    }
    \label{fig:framework}
    \vspace{-10pt}
\end{figure*}

\vspace{3pt}
\noindent\textbf{Image Blending} In video frame interpolation and image stitching, image blending is an important task which aims at blending two images with overlapping regions together and form a realistic image.
The architecture of blending modules used in video frame interpolation methods can be divided into Unet~\cite{unet}-based~\cite{enhanced, Space-time} and GridNet~\cite{GridNet}-based structures~\cite{Context, soft-splat, asymmetric}, and both methods produce realistic synthesized images. 
Among these two architectures, we choose to modify UNet and devise a Siamese encoder-decoder architecture with skip-connections for the blending module.

Several stitching studies specifically focus on handling seams in image blending. 
In order to encourage smooth transitions along the seam, cost functions in gradient domain~\cite{gradient} or seam costs~\cite{largemotion} are defined.
Other methods use seam-cutting to align the images and employ blending techniques such as local seam blending~\cite{DHW}, multi-band blending algorithm~\cite{ptis}, or iteratively improve seam quality~\cite{seagull}.
Along with the development of deep-learning,~\cite{view-free} was the first to propose a fully deep-learning-based framework for image stitching, especially targeting view-free stitching.
To reduce unwanted seams, several attempts such as an edge-preserved deformation branch~\cite{edge-preserved}, content loss~\cite{view-free} and seam loss~\cite{uis} have been made.
Contrary to the above methods, we propose SIGMo, which utilizes adversarial loss to eliminate unnatural seams and misalignments in the final stitched results.

\vspace{3pt}
\noindent\textbf{Hole Filling}
Among computer vision tasks that require hole filling, the most actively researched field is image inpainting. 
For deep image inpainting, various techniques have been proposed to concentrate on filling holes with relevant texture and structure information.
These include using partial convolutions conditioned only on valid pixels, utilizing learnable bidirection attention maps, or using a mutual encoder-decoder with feature equalization~\cite{inpaint-bidir, inpaint-partial, inpaint-mut}.
Our proposed method uses forward warping when warping the target image to the reference image domain, which inevitably leads to holes in the warped image.
The main difference between the previously mentioned deep image inpainting methods and our proposed method is that our approach has no explicit ground truth hole mask.
To address this problem, we devise hole-aware reconstruction loss and adversarial loss to eliminate holes in the final stitched result.

\section{Pixel-wise Deep Image Stitching}
For image stitching, we propose to directly estimate the pixel-wise warp field instead of estimating the homography-based warp. 
To achieve this goal, we propose a novel pixel-wise deep image stitching framework which will be detailed in this section.
Through this approach, we can effectively perform image stitching in more challenging conditions especially the scenes involving large parallax or non-planar regions. 

\subsection{Overall Framework}
The proposed framework consists of two modules: Pixel-wise Warping Module (PWM) and Stitched Image Generating Module (SIGMo), as shown in Fig.~\ref{fig:framework}. Since our goal is to stitch the target image into the reference image plane in a pixel-wise manner, we first reposition each pixel in the target image to obtain the warped target image using PWM. Then, the warped target image from PWM and the reference image are fed into the SIGMo to generate the final stitched image.

In detail, given pair of reference and target images $(I^R, I^T)$, PWM first estimates the pixel-wise warp field $W_{T \rightarrow R}$ from $I^T$ to $I^R$ including both overlap and non-overlap region. Then, from the estimated warp field, we obtain the warped target image $I^{WT}$ by relocating each pixel in $I^T$ into the $I^R$ domain using differentiable forward warping~\cite{soft-splat}. Since the forward warping process inevitably creates undesirable holes in the warped target image, this side effect should be eliminated in the subsequent procedure for a plausible stitching image. Therefore, we design the SIGMo to fill the holes and generate the final stitched image $I^S$, while naturally blending $I^{WT}$ and $I^R$ to reduce the undesirable artifacts including seams and misalignments.

\begin{figure*}[t]
    \centering
    \includegraphics[width=0.99\linewidth]{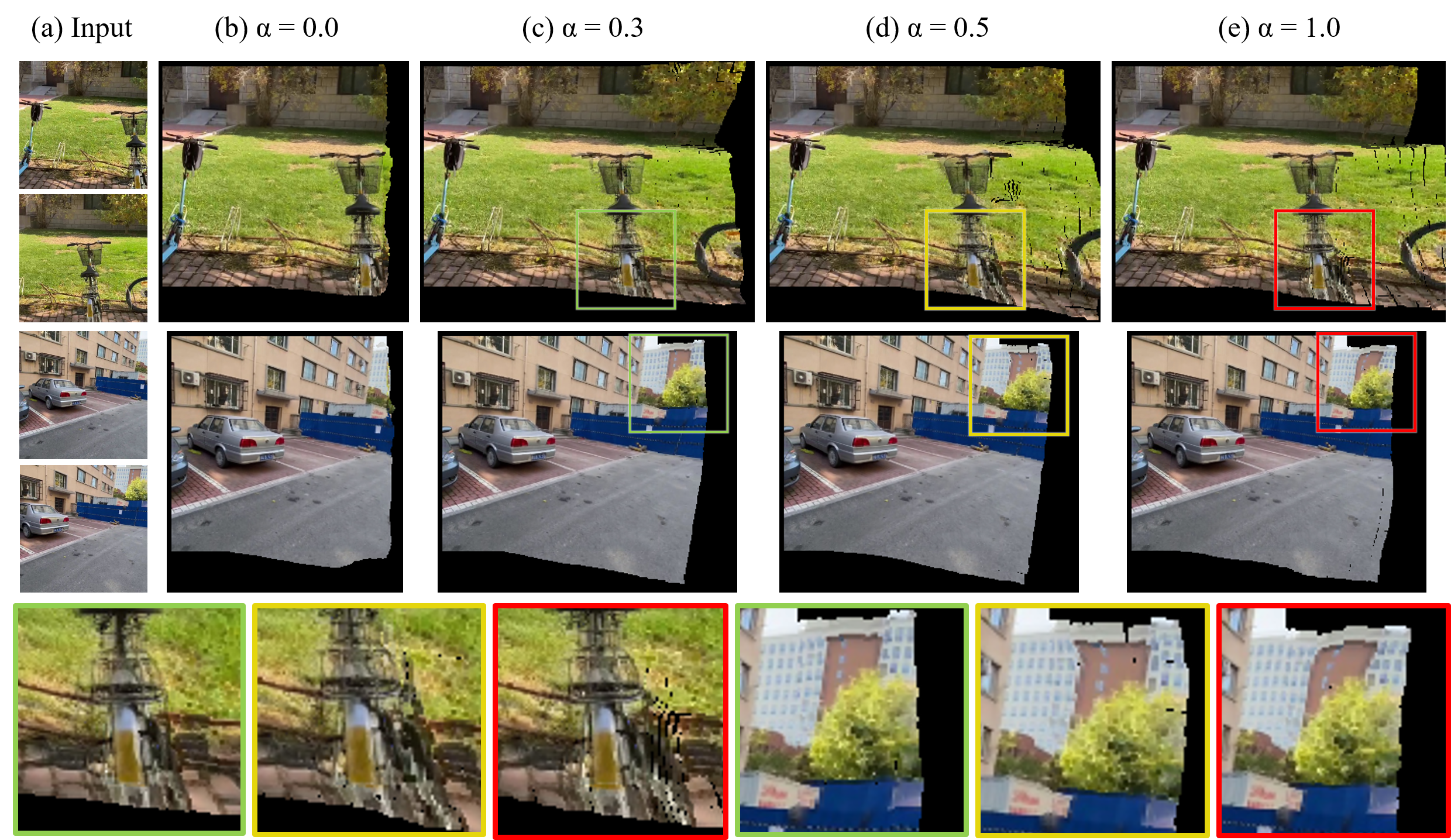}
    \vspace{-10pt}
    \caption{Qualitative comparisons of the proposed PWM with different NOV supervision weights.
    The stitched image is obtained by averaging the overlap regions of reference and warped target images.
    Input reference (upper) and target (lower) images are in (a). The stitched images of PWM are in (b)-(e), and the $\alpha$ is 0.0, 0.3, 0.5, 1.0, respectively.}
    \label{fig:raft-ablation}
    \vspace{-10pt}
\end{figure*}

\subsection{Pixel-wise Warping Module (PWM)}

Instead of estimating the homography-based regional warp, we propose a method to estimate the pixel-wise warp field directly.
Our approach has similarities with OF estimation in the nature of the task.
Especially on the OV region between two images, the proposed scheme and the OF estimation have an identical purpose: estimating pixel-wise 2D warp from one image to the other image based on point correspondences.
Therefore, we borrow the recent network architectures from the researches in the field of OF estimation~\cite{flownet2, raft}.
With ground truth supervision of the proposed dataset, we can train the model and obtain the reliable pixel-wise warp field of the OV region.

But simply imposing OF-like losses on the existing OF estimation models cannot be a complete solution for warp estimation in the perspective of image stitching.
Unlike estimating the warp on the OV region, predicting the pixel-wise warp field of the NOV region is an ill-defined problem due to the absence of correspondences.
Providing GT supervision for the NOV region can work as an approximated guidance since existing OF models can learn the relationship between OV and NOV due to the wide receptive field.
However, excessively forcing the model to reduce the loss between the predicted warp and GT warp on the NOV region may lead to overfitting on the trained dataset, similar to that of the monocular depth estimation task.
Moreover, for the NOV region, in the perspective of image stitching, the warp field that leads to a plausible stitching result is not unique.

For this reason, instead of applying an identical loss function for all pixels on the image, we regularize the loss for the pixels on the NOV region by weighting.
The proposed warp loss for PWM is defined as Eq. (\ref{eq:loss-pwm}).
\begin{equation}
\label{eq:loss-pwm}
    \mathcal{L}_{PWM} = \parallel W_{ov} - W^{gt}_{ov} \parallel_{1} + \alpha \parallel W_{nov} - W^{gt}_{nov} \parallel_{1}
\end{equation}
where $W_{ov}$ and $W_{nov}$ are the estimated warp field in overlap and non-overlap region, respectively. $W^{gt}_{ov}$ and $W^{gt}_{nov}$ are the GT labels for $W_{ov}$ and $W_{nov}$. 
Here, to make the model mainly focus on the OV region while keeping minimum guidance on the NOV region, $\alpha$ regularizes the non-overlap warp supervision.
To know which region a certain pixel belongs to, we use a mask for overlap region in the proposed dataset, where the process of acquiring the mask will be detailed in \textit{Appendix}.

To find a proper degree of regularization for NOV supervision, we conduct an experiment on the proposed dataset while adjusting $\alpha$ in Eq.~\ref{eq:loss-pwm}.
Fig.~\ref{fig:raft-ablation} visualizes qualitative comparisons of the proposed PMW with different $\alpha$.
Imposing GT only on the OV region ($\alpha=0$) completely fails to predict pixel-wise warp on the NOV region, which is what we expected.
With $\alpha>0$, PWM succeeds to make a reasonable prediction for the NOV region.
We observe that as we increase $\alpha$, the warped target images show a tendency to be distorted on both the OV and NOV region, due to the over-fitting that we were concerned about.

We attempted to apply an epipolar loss based on Sampson distance error, which constrains the warped pixel to be on the epipolar line of the pixel before being warped.
This was intended to provide guidance for estimating the pixel-level warp on the NOV region.
However, unlike the expectation, applying the epipolar loss did not bring any qualitative or quantitative improvements.
Moreover, we also observe that the epipolar loss decreases while training PWM only with the loss function in Eq.~\ref{eq:loss-pwm}.
It seems that the act of applying GT supervision on NOV regions already implicitly provides the information about the epipolar constraint.
Additional discussion about the epipolar loss will be detailed in \textit{Appendix}.

After the pixel-wise warp field $W_{T \rightarrow R}$ is obtained using ground truth supervision, we transform each pixel in $I^T$ using $W_{T \rightarrow R}$ in order to obtain the warped target image $I^{WT}$.
For transforming each pixel, we use softmax splatting~\cite{soft-splat}, which is a well-known forward warping method in video frame interpolation.

\begin{figure*}[ht]
    \centering
    \includegraphics[width=0.99\linewidth]{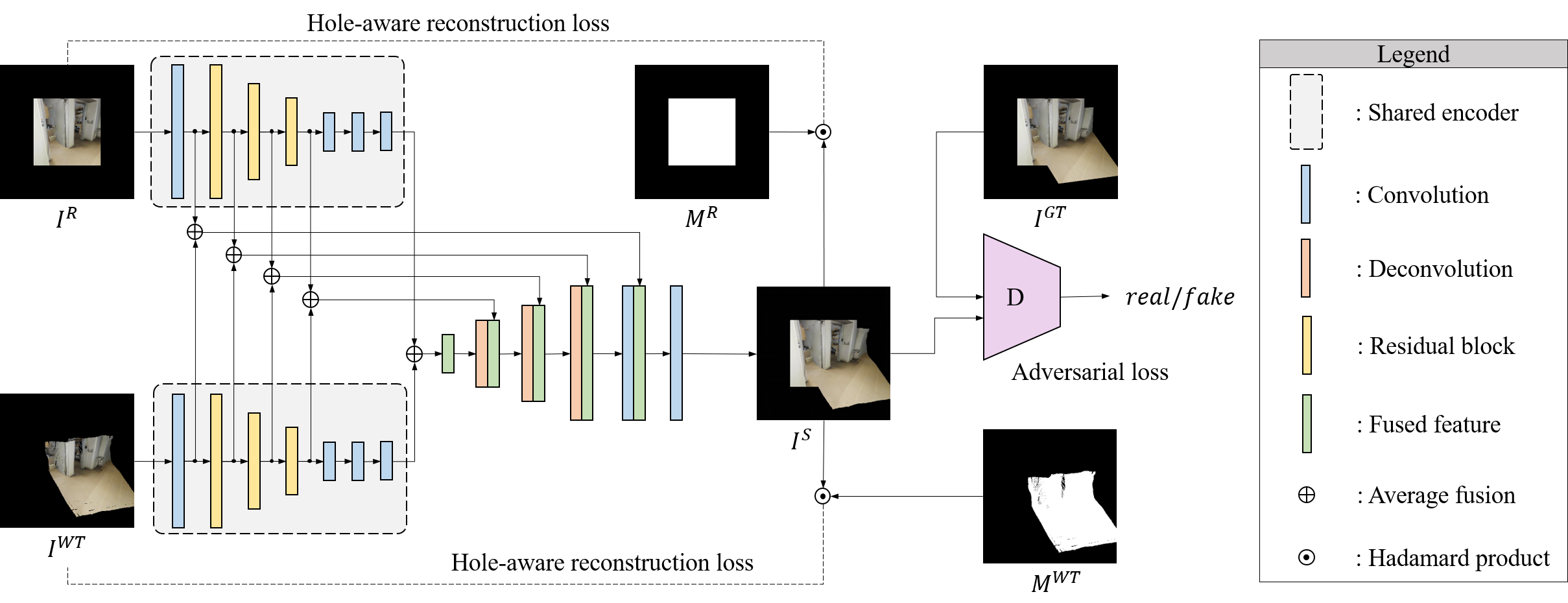}
    \vspace{-10pt}
    \caption{Network architecture of SIGMo. $I^R$ and $I^{WT}$ are fed into the shared encoders and fused with addition. Then, the decoder takes the fused features as input to obtain a final stitched image $I^S$. In the decoding phase, separate intermediate features of $I^{R}$ and $I^{WT}$ are fused and used as skip connections. To handle the undesirable artifacts while preserving the structures of input images, hole-aware reconstruction loss and adversarial loss are devised.}
    \label{fig:SIGMo}
    \vspace{-10pt}
\end{figure*}

\subsection{Stitched Image Generating Module (SIGMo)}
We design SIGMo to generate a plausible stitched image  $I^{S}$ with given reference image $I^R$ and warped target image $I^{WT}$ as in Fig.~\ref{fig:SIGMo}. Specifically, SIGMo handles three undesired artifacts: 1) holes, 2) misalignments, and 3) seams. 
Since the forward warping process in PWM creates unwanted holes in $I^{WT}$, we need to fill the holes for plausible image stitching. 
A simple way to resolve this problem is to adapt image inpainting techniques, but these methods cannot distinguish between holes and background.
Distinguishing these holes from the background is complicated, which makes it hard to interpolate neighboring pixel values to fill the holes. 
In addition, due to the errors in $W_{T \rightarrow R}$ and color differences between $I^R$ and $I^{WT}$, misalignments and seams might occur, respectively. 

To alleviate these artifacts, we devise hole-aware reconstruction loss and adversarial loss. In particular, we want SIGMo to remove the artifacts while preserving the overall structures of input $I^R$ and $I^{WT}$. 
Therefore, when adversarial loss reduces the artifacts, hole-aware reconstruction loss is used to constrain SIGMo and preserve the input images.

For hole-aware reconstruction loss, we first obtain reference occupancy mask $M^{R}$ and warped target occupancy mask $M^{WT}$. Since the size of $I^{WT}$ can be larger than $I^{R}$, we predefine the size of the resulting stitched image $I^{S}$ to be twice the size of $I^R$. We pad or crop $I^{R}$ and $I^{WT}$ in order to match the size with $I^{S}$. For convenience, from now on, $I^R$ and $I^{WT}$ also stand for the padded or cropped $I^R$ and $I^{WT}$, respectively. Then, we can easily obtain $M^{R}$ or $M^{WT}$ with a value of $1$ for where the actual pixel occupies in $I^{R}$ or $I^{WT}$ and $0$ otherwise, respectively. The hole-aware reconstruction loss is then defined as Eq. (\ref{eq:loss-recon}).

\begin{equation}
\begin{aligned}
\label{eq:loss-recon}
    \mathcal{L}_{recon} &= \mathcal{L}^{R}_{recon} + \mathcal{L}^{WT}_{recon} \\
    \mathcal{L}^{R}_{recon} &= \parallel I^{S} \odot M^{R} - I^{R} \parallel_{1} \\
    \mathcal{L}^{WT}_{recon} &= \parallel I^{S} \odot M^{WT} - I^{WT} \parallel_{1}
\end{aligned}
\end{equation}

For adversarial loss, we use an unconditional patch discriminator $D$ which determines whether or not the small patches of $I^{S}$ are real~\cite{pix2pix}. Here, a sample GT image of the proposed dataset $I^{GT}$ is used for a real image. The objective function based-on the LSGAN loss~\cite{lsgan} can be represented as Eq. (\ref{eq:loss-adv}):
\begin{equation}
    \label{eq:loss-adv}
    \begin{aligned}
    \mathcal{L}_{D} &= \mathbb{E}_{I^{GT}}[(D(I^{GT})-1)^{2}] + \mathbb{E}_{I^{S}}[(D(I^{S}))^{2}] \\
    \mathcal{L}_{adv} &= \mathbb{E}_{I^{S}}[(D(I^{S})-1)^{2}]  
    \end{aligned}
\end{equation}

The total loss function of SIGMo is as follows:

\begin{equation}
    \vspace{-3pt}\mathcal{L}_{SIGMo} = \lambda_{r}\mathcal{L}_{recon} + \lambda_{a}\mathcal{L}_{adv}
\end{equation}
where $\lambda_{r}$ and $\lambda_{a}$ are weighting parameters of $\mathcal{L}_{recon}$ and $\mathcal{L}_{adv}$, respectively.

We devise a Siamese encoder-decoder architecture with skip-connections for SIGMo. Instead of fusing $I^R$ and $I^{WT}$ at an early stage, we fuse them after feeding them into the shared encoder separately, as in Fig.~\ref{fig:SIGMo}. During the decoding phase, separate intermediate features of $I^{R}$ and $I^{WT}$ are fused and used as skip connections. We use average fusion for the fusing method.

\begin{figure*}[ht]
    \centering
    \includegraphics[width=0.99\linewidth]{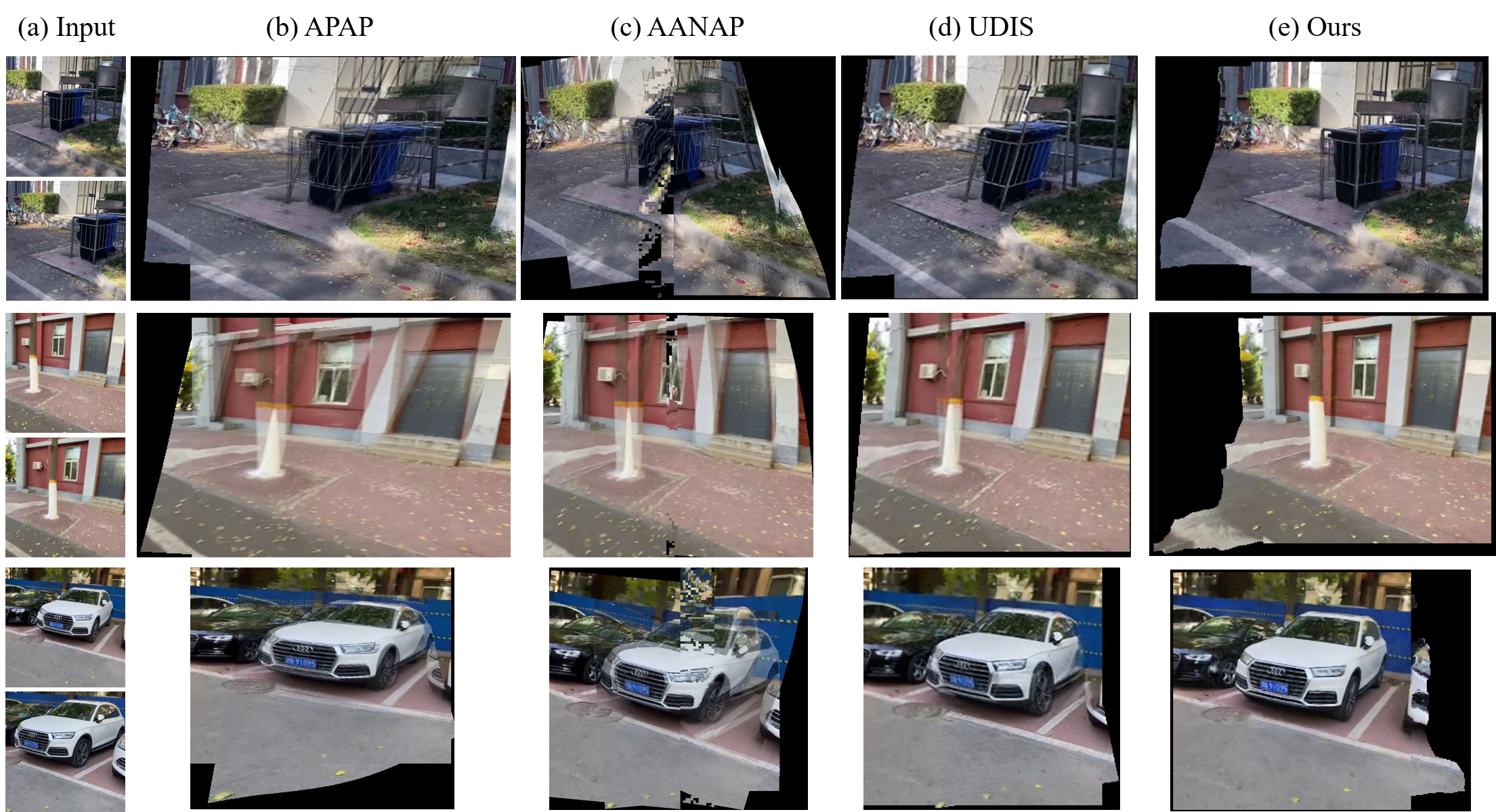}
    \vspace{-8pt}
    \caption{Qualitative comparisons of the stitched images of various methods.
    From left to right: (a) input images (reference $\uparrow$, target $\downarrow$), (b) APAP~\cite{apap}, (c) AANAP~\cite{aanap}, (d) UDIS~\cite{uis}, (e) our proposed method.}
    \label{fig:comparison}
    \vspace{-10pt}
\end{figure*}

\begin{figure*}[ht]
    \centering
    \includegraphics[width=0.99\linewidth]{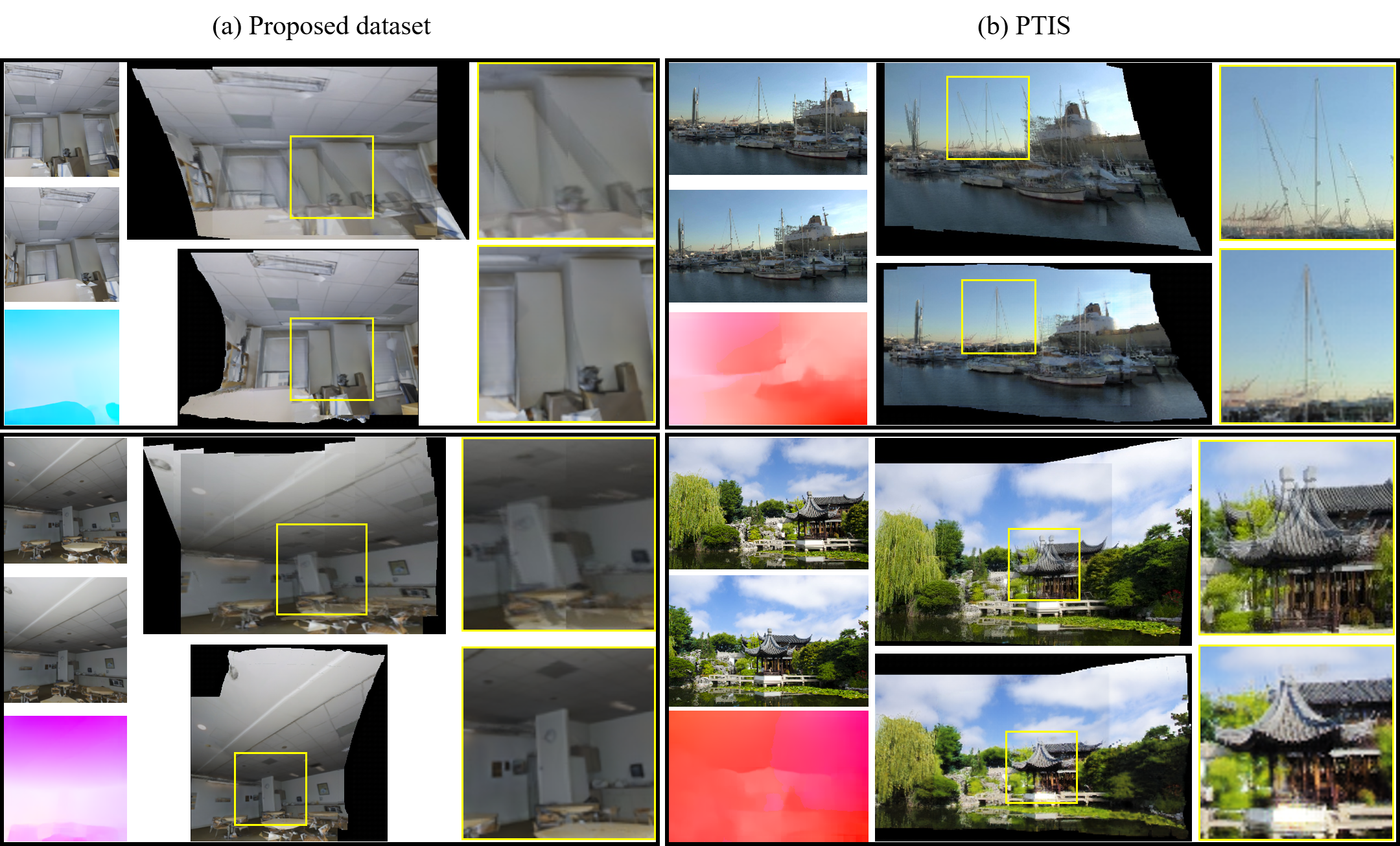}
    \vspace{-5pt}
    \caption{Qualitative comparisons on large parallax images. From left to right for each block, first column: reference image, target image, flow between two images from our PWM, second column: the results of APAP~\cite{apap} and our proposed method, third column : enlarged image patches of the results.}
    \label{fig:lapis}
     \vspace{-10pt}
\end{figure*}

\vspace{-5pt}
\section{Experiments}
\vspace{-2pt}
\subsection{Proposed Dataset} \label{dataset}
\vspace{-3pt}
The largest obstacle to introduce learning-based approaches to the field of image stitching is the absence of a large-scale dataset.
Several existing studies published datasets for qualitative comparison, however, the datasets are relatively small in terms of size and do not include GT warp between the images, even in the OV region.
A recent approach based on unsupervised learning proposes a large-scale dataset that contains the frames collected from the videos~\cite{uis}.
Nevertheless, this dataset also does not include GT correspondences, which is insufficient to train the proposed framework based on pixel-wise warp estimation.

So, to train and evaluate the proposed framework, we utilize a virtual environment~\cite{stanford2d3d-dataset} (Apache License 2.0) to build a novel large-scale dataset for image stitching.
From multiple scenes with virtual cameras at two different views, we render image pairs: the reference image $I^R$ and the target image $I_T$. 
To mimic situations where image stitching is generally applied, we make a pair of rendered images share some overlap regions.
Our proposed dataset contains 4,098 pairs of images for training and 919 pairs for test.

With the projective camera matrices of the virtual cameras, ground truth warps are obtained with respect to the reference image.
As a sample, the proposed dataset also provides stitching results constructed from the reference image and the target image warped according to the GT warp.
Although the sample result is constructed with the GT warp, estimating the warp for NOV region is an ill-defined problem, and therefore is just one of the possible realizations in the perspective of image stitching.
More details about our dataset can be found in \textit{Appendix}

\subsection{Experimental Setup}
We train and evaluate our method on our proposed dataset, and for comparison, additionally use datasets proposed in~\cite{ptis, uis}.
Although we resized the input images to 224$\times$224 for training our warp estimation network, our model has no explicit restriction of input size. 
We train our network for 500 epochs with batch size of 8. 
All modules, PWM and SIGMo, are optimized with AdamW optimizer~\cite{adamw} with $\beta_{1}$ = 0.5, $\beta_{2}$ = 0.999, and learning rate 1$\times10^{-4}$.
We use $\alpha$ = 0.3, $\lambda_{r}$ = 1, and $\lambda_{a}$ = 0.1 for training.

Among various off-the-shelf optical flow estimation models, we choose to use RAFT~\cite{raft} architecture, which has shown promising results in optical flow estimation, as the warp estimation network in PWM.
Note that any optical flow estimation model can be used as the warp estimation network, and the qualitative results will be discussed in the \textit{Appendix}.
\vspace{-5pt}

\vspace{5pt}
\subsection{Experimental Results}

\noindent\textbf{Comparisons with Existing Methods} 
Fig.~\ref{fig:comparison} shows the qualitative comparison between the proposed method and the previous image stitching methods~\cite{apap, aanap, uis}.
The input images are from the dataset proposed in UDIS~\cite{uis}.
As our intentions, the proposed method robustly stitches the input images without misalignments and severe distortion, while the other methods often fail at converging or produce serious misalignments.
More comparative experiments can be found in the \textit{Appendix}.

\vspace{3pt}
\noindent\textbf{Challenging Scenes} 
We illustrate the qualitative comparisons of our method with APAP~\cite{apap} on challenging scenes with large parallax in Fig.~\ref{fig:lapis}. 
The input image pairs are provided by the proposed dataset and the dataset proposed in \cite{ptis}.
The results show that our method based on pixel-wise warping is robust on large parallax scenes, whereas APAP fails in both cases.
Since APAP uses local regional warp based on homography, it is bound to fail at scenes with curvature or large parallax.
On the other hand, our model can obtain robust and plausible stitching results without misalignments, even for scenes with large baselines and dramatic scene depth differences.

\begin{figure}[t]
    \centering
    \includegraphics[width=0.99\linewidth]{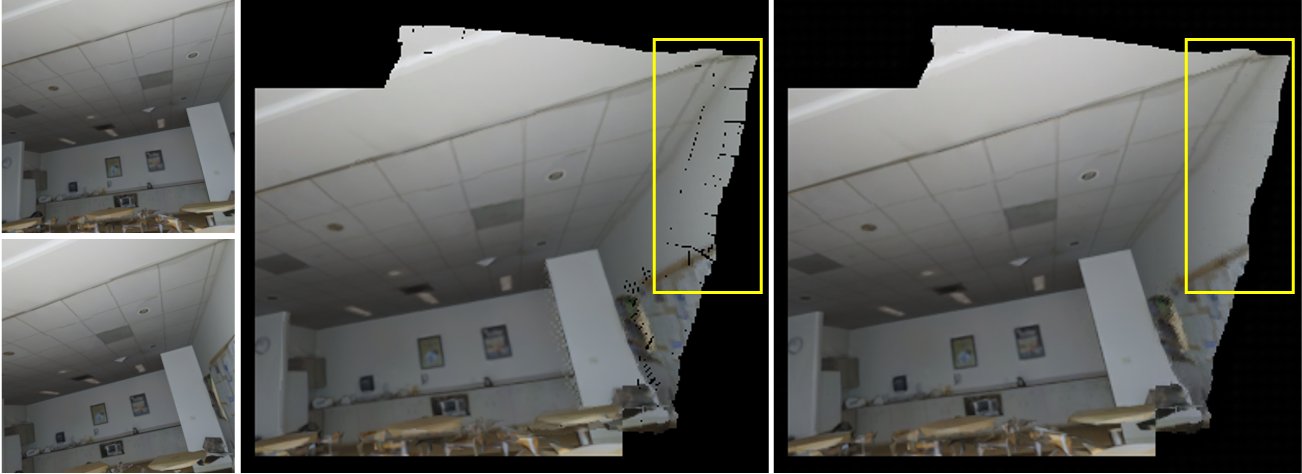}
    \vspace{-7pt}
    \caption{Visualization of hole filling with the proposed SIGMo. From left to right: The input image pair from the proposed dataset, the result of average blending, and the result of SIGMo.}
    \vspace{-15pt}
    \label{fig:hole}
\end{figure}

\vspace{3pt}
\noindent\textbf{Hole Filling} 
We visualize the hole filling ability of the proposed SIGMo in Fig.~\ref{fig:hole}. We compare the result of SIGMo with the average blending method, which obtained by averaging the overlap regions of reference and warped target images. As shown in the yellow boxes, with the help of hole-aware reconstruction loss and adversarial loss, SIGMo successfully handles the undesirable holes in the warped target image, caused by the forward warping in PWM.

\vspace{-5pt}
\section{Limitations}
\begin{figure}[t]
    \centering
    \includegraphics[width=0.99\linewidth]{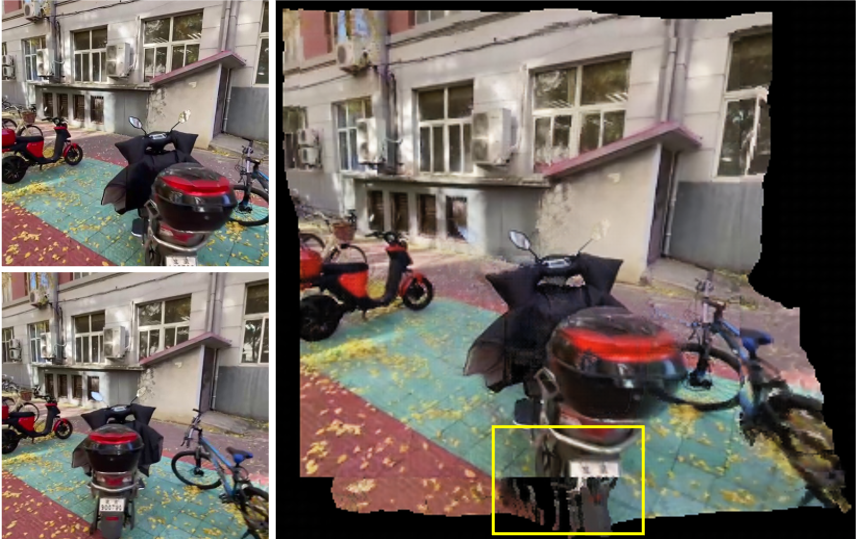}
    \vspace{-7pt}
    \caption{Failure case of our method. In the yellow box, we can observe misalignment and holes due to the mis-predicted warp and the lack of information. The input image pair is from the dataset published by~\cite{uis}.}
    \label{fig:failure}
    \vspace{-15pt}
\end{figure}

\vspace{-5pt}
The yellow boxes in Fig.~\ref{fig:failure} show the main limitations of the proposed method: hole and mis-predicted warp, especially in the NOV region. 
We make SIGMo generate plausible stitching results with the help of hole-aware reconstruction loss and adversarial loss. However, it is still difficult to fill the large hole caused by occlusion or lack of information.
Also, though imposing GT supervision on the NOV region shows some degree of success, the ill-defined nature of estimating pixel-wise warp on the NOV region is not fully addressed by the proposed framework in this paper.
Finally, due to the limited receptive field of the proposed DL-based framework and memory issue, our method can only handle images that have a resolution in the limited range.

\vspace{-5pt}
\section{Conclusion}
\vspace{-5pt}

Most of the existing studies in image stitching have exploited the homography-based warp, which causes parallax problems.
In this paper, to address the large parallax problem, we propose to estimate the pixel-wise warp instead of the homography-based warp.
For this, we propose a novel framework composed of the Pixel-wise Warping Module (PWM) and the Stitched Image Generating Module (SIGMo).
The PWM estimates and applies the warp from the target image to the reference image.
To handle the ill-defined nature of the estimating warp for NOV regions, we propose an NOV-regularized loss and conduct an experiment to find a proper degree of regularization.
Then, with the help of the proposed hole-aware reconstruction loss and the adversarial loss, SIGMo generates plausible stitching results by combining the reference image and the warped target image.
To train the proposed framework, we also build a large-scale synthetic dataset for image stitching.
As a result, with the proposed framework and the dataset, we obtain outperforming stitching results in terms of stitching quality.
Our method can handle challenging scenes with large parallax as we intended, where the other methods usually fail.
It supports the superiority of the proposed pixel-wise warp estimation approach compared to the homography-based approaches in the field of image stitching.
Though the proposed method has a limitation in resolution due to a memory issue, we show that our approach is a promising direction for image stitching.
It will be interesting for future research to improve the proposed method so that it can be applied at a higher resolution.

{\small
\bibliographystyle{ieee_fullname}
\bibliography{egbib}
}

\newpage
\begin{appendix}

\begin{figure}[t]
    \centering
    \includegraphics[width=0.99\linewidth]{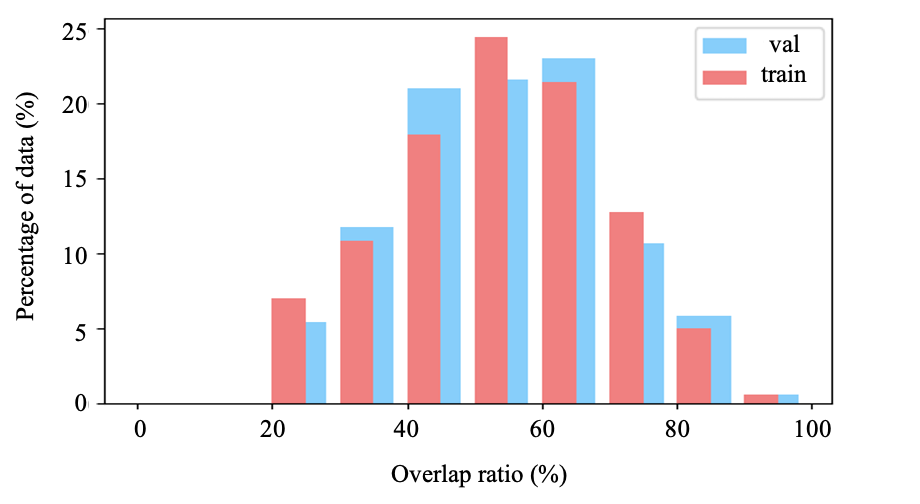}
    \vspace{-7pt}
    \caption{Data distribution of train and validation set of the proposed dataset with respect to the overlap ratio of reference and target images.}
    \label{fig:dataset-histogram}
    \vspace{-10pt}
\end{figure}
%

\begin{figure}[t]
    \centering
    \includegraphics[width=0.99\linewidth]{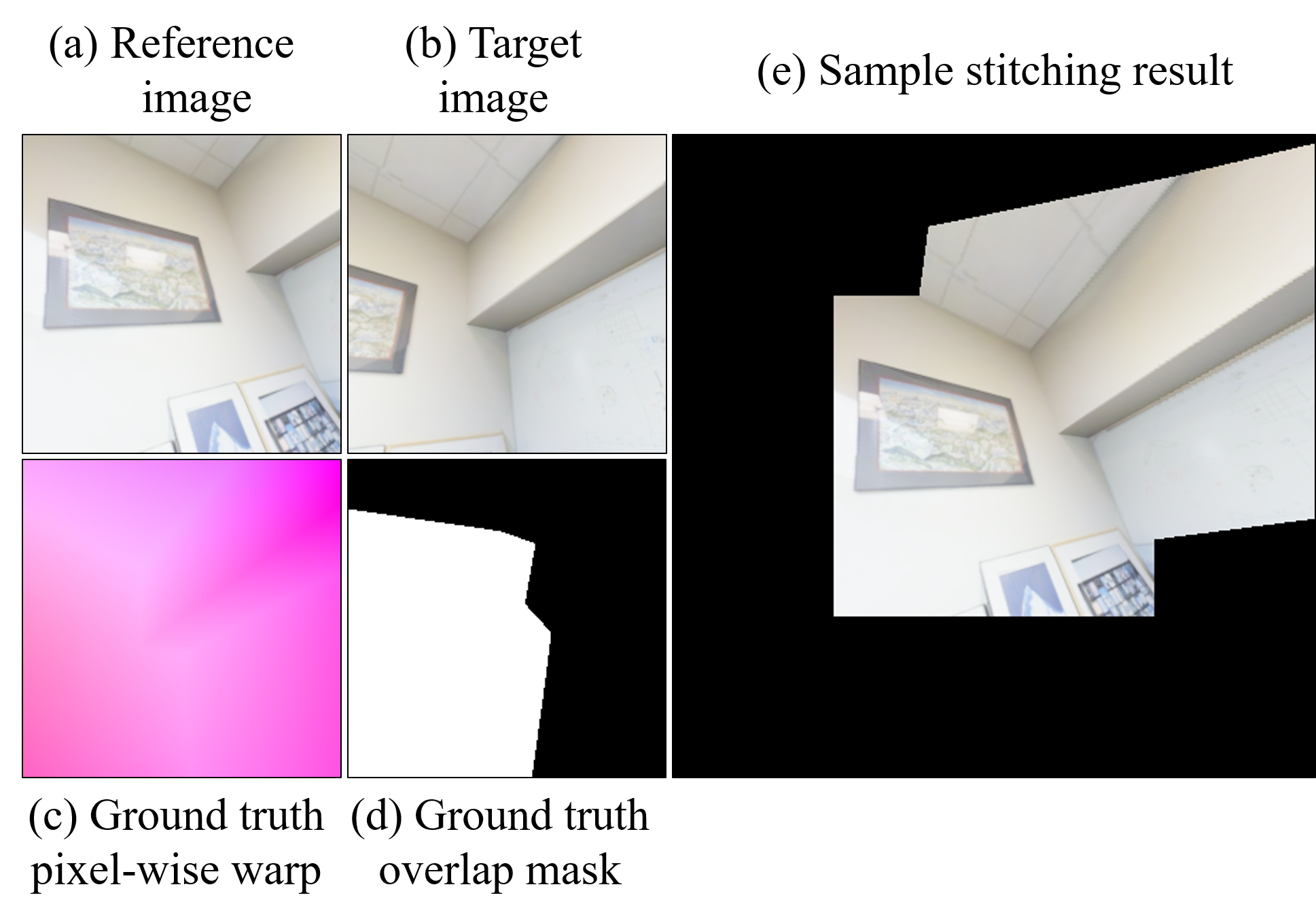}
    \vspace{-7pt}
    \caption{Sample of the proposed dataset. From (a) to (e): reference image, target image, GT pixel-wise warp, GT overlap mask, and sample stitching result according to the GT pixel-wise warp.}
    \vspace{-15pt}
    \label{fig:dataset-sample}
\end{figure}
\section{Dataset Details}
To train and evaluate the proposed framework, we build a novel large-scale dataset for image stitching.
For this, we utilize a virtual environment~\cite{stanford2d3d-dataset} (Apache License 2.0).
The environment is composed of 3D meshes and textures reconstructed from RGB images and corresponding depth, and thereby the scenes are photorealistic compared to the fully virtual world.
We render image pairs from multiple scenes with virtual cameras at two different views: the reference image and the target image. 
To mimic situations where image stitching is generally applied, we make a pair of rendered images share overlap regions.
The data distribution of the proposed dataset with respect to the overlap ratio between reference and target images is shown in Fig.~\ref{fig:dataset-histogram}.

Our proposed dataset contains 4,098 pairs of images for training and 919 pairs for the test.
With the projective camera matrices of the virtual cameras, ground truth warps are obtained with respect to the reference image.
As a sample, the proposed dataset also provides stitching results constructed from the reference image and the target image warped according to the GT warp.
Although the sample result is constructed with the GT warp, estimating the warp for the NOV region is an ill-defined problem, and therefore is just one of the possible realizations in the perspective of image stitching.
The sample of the proposed dataset is shown in Fig.~\ref{fig:dataset-sample}.

\begin{figure}[t]
    \centering
    \includegraphics[width=0.99\linewidth]{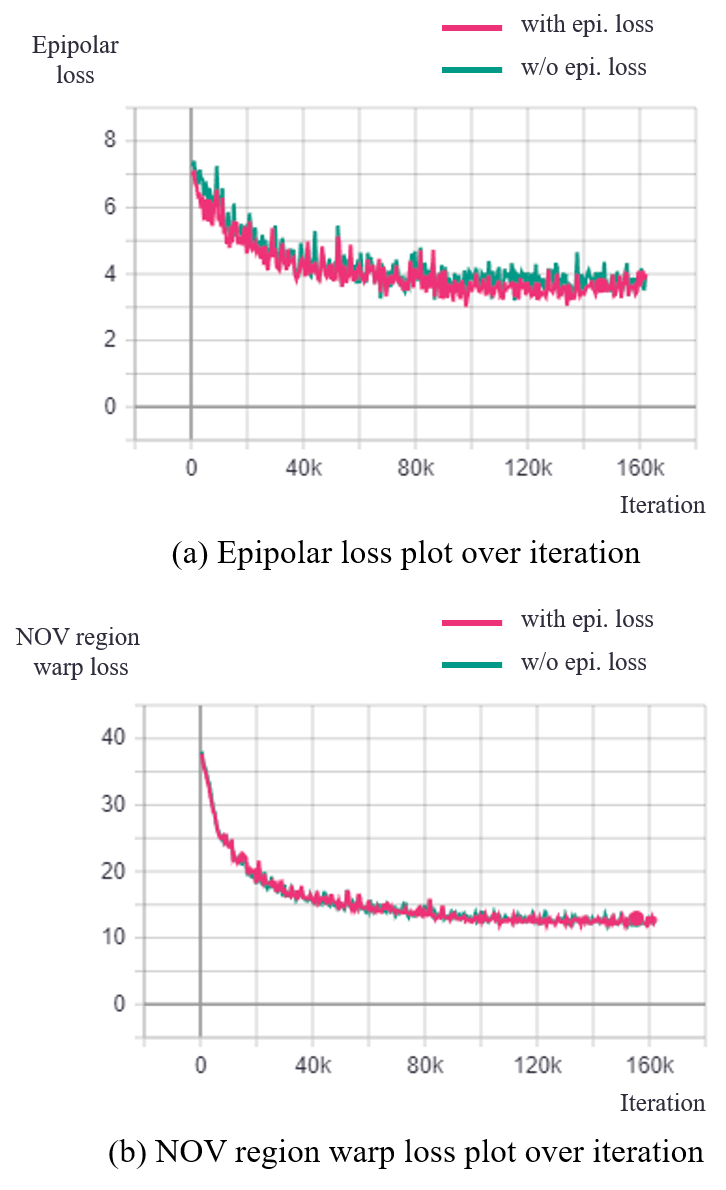}
    \vspace{-7pt}
    \caption{Qualitative comparisons between the model trained with epipolar loss and the model without epipolar loss. (a): Epipolar loss plot over iteration. (b): Warp loss plot over iteration. Both models are trained and evaluated on the train and validation set of the proposed dataset, respectively.}
    \vspace{-20pt}
    \label{fig:epipolar_plot}
\end{figure}

\section{Discussion on Epipolar Loss}
As we mentioned in the main paper, we attempted to provide an additional guideline based on epipolar geometry for PWM to estimate the pixel-wise warp on the NOV region.
We devise a loss function that has the form of Sampson distance error to constrain the warped pixel to be on the epipolar line of the pixel before being warped.
For this, we use GT fundamental matrix $F$ computed from the projective matrices of the virtual cameras.
Formally, for all pixels $\{\mathbf{x}_{i}\}^{N}_{1}$ in target image $I^T$, we can define the corresponding warped pixel $\{\mathbf{x}^{'}_{i}\}^{N}_{1}$ in $I^R$ domain using the estimated warp field $W_{T \rightarrow R}$ of PWM. For clarity, we will denote warp field $W_{T \rightarrow R}$ as $W$. The epipolar loss in PWM is defined as Eq. (\ref{eq:loss-epi}). Here, $(\mathbf{F}\mathbf{x}_{i})_k$ represents $k$-th entry of the vector $\mathbf{F}\mathbf{x}_{i}$.

\begin{equation}
\label{eq:loss-epi}
    \mathcal{L}_{epi} = \sum_{i=1}^{N}\frac{(\mathbf{x}^{'T}_{i}\mathbf{F}\mathbf{x}_{i})^2}{(\mathbf{F}\mathbf{x}_{i})^{2}_{1} + (\mathbf{F}\mathbf{x}_{i})^{2}_{2} + (\mathbf{F}^{T}\mathbf{x}^{'}_{i})^{2}_{1} + (\mathbf{F}^{T}\mathbf{x}^{'}_{i})^{2}_{2}}
\end{equation}

However, unlike the expectation, applying the epipolar loss did not bring any improvements.
Fig.~\ref{fig:epipolar_plot} shows that the epipolar loss  decreases while training PWM only with the loss function in Eq.~(\textcolor{red}{1}) in the main paper.
Moreover, when we compute the loss between the estimated pixel-wise warp and the GT pixel-wise warp on the NOV region, the models show an almost similar tendency regardless of whether we use epipolar loss or not.
We interpret this result as that the act of applying GT supervision on NOV regions already implicitly provides information about the epipolar constraint.
It would be an interesting future research to utilize the epipolar loss for training PWM in a semi-supervised or unsupervised manner.
In this case, instead of using GT, the fundamental matrix can be estimated from the obtained correspondences on the OV region. 

\section{Adopting Different Optical Flow Model}
We propose to estimate the pixel-wise warp from target image to reference image.
For this, as a warp estimation network in PWM, we employ the network architecture from the researches in the field of optical flow(OF) estimation.
Experiments in the main paper are conducted with RAFT~\cite{raft}, which has shown substantial results for OF estimation.

\begin{figure*}[t]
    \centering
    \includegraphics[width=0.99\linewidth]{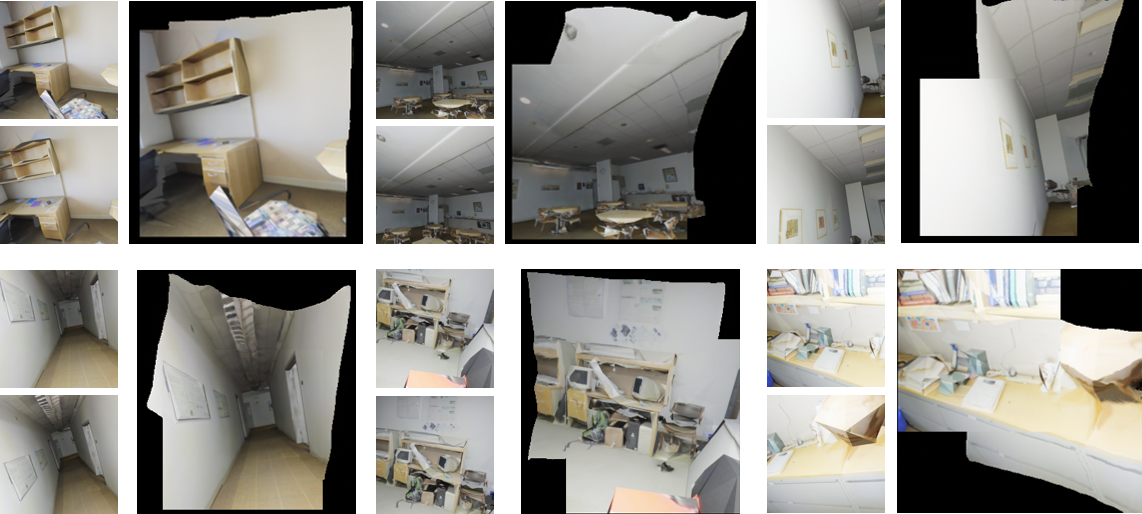}
    \caption{Qualitative results using FlowNet2.0~\cite{flownet2} as the warp estimation module on our proposed dataset.}
    \label{fig:flownet-supple}
\end{figure*}

In this section, we show the qualitative results of the proposed framework with FlowNet2.0~\cite{flownet2} in Fig.~\ref{fig:flownet-supple}.
The results show that our framework achieves promising image stitching results regardless of which OF estimation architecture is used. 
Moreover, we can expect that the performance of the proposed framework can be improved by using a better OF estimation architecture. 

\begin{table}[t]
\small
\begin{center}
\caption{Pixel-wise end-point-error of PWM in the overlap (OV) and non-overlap (NOV) regions according to different overlap ratios and NOV-regularization parameter $\alpha$. \textbf{Bold} numbers represent the best results, while \underline{underlined} numbers are the second-best ones.}
\begin{tabular}{cc||ccc||c}
    \hline
    \multicolumn{2}{c||}{\multirow{2}{*}{\begin{tabular}[c]{@{}c@{}}$\alpha$\end{tabular}}} & \multicolumn{3}{c||}{Overlap ratio (\%)} & \multirow{2}{*}{Total mean} \\ 
    \multicolumn{2}{c||}{}                                                                                            & 20$\sim$40       & 40$\sim$60       & 60$\sim$80      &                        \\ \hline\hline
    \multicolumn{1}{c|}{\multirow{2}{*}{0.0}}                                  & OV                                  & 8.803       & 3.403       & 2.835      & \textbf{4.123}                  \\
    \multicolumn{1}{c|}{}                                                      & NOV                                 & 76.293      & 50.546      & 25.714     & 44.315                 \\ \hline
    \multicolumn{1}{c|}{\multirow{2}{*}{0.3}}                                  & OV                                  & 13.572      & 4.091       & 3.269      & \underline{5.432}                  \\
    \multicolumn{1}{c|}{}                                                      & NOV                                 & 32.313      & 15.736      & 8.818      & \textbf{15.799}                 \\ \hline
    \multicolumn{1}{c|}{\multirow{2}{*}{0.5}}                                  & OV                                  & 17.142      & 5.054       & 3.487      & 6.576                  \\
    \multicolumn{1}{c|}{}                                                      & NOV                                 & 37.349      & 17.749      & 8.793      & \underline{17.576}                 \\ \hline
    \multicolumn{1}{c|}{\multirow{2}{*}{1.0}}                                  & OV                                  & 16.844      & 6.259       & 4.065      & 7.291                  \\
    \multicolumn{1}{c|}{}                                                      & NOV                                 & 34.538      & 18.132      & 9.444      & \underline{17.576}                 \\ \hline
\end{tabular}%
\label{table:regularization}
\end{center}
\end{table}
\section{Regularization Parameter in PWM}

In addition to qualitative comparisons in the main paper, we also conduct the quantitative comparison on different regularization parameter $\alpha$ for reference as shown in Table~\ref{table:regularization}. Although the low end-point-error of the estimated warp does not guarantee plausible stitching results, we can regard this metric as acceptable guidance. 
As expected, imposing GT warp supervision only on the OV region ($\alpha=0$) cannot help the model predict a reasonable warp on the NOV region.
With a tendency similar to qualitative results, the pixel-wise warp on both OV and NOV regions is quantitatively degraded as $\alpha$ increases.
Therefore, in addition to the qualitative comparison (Fig.~\textcolor{red}{3}) in the main paper, the quantitative results also support that imposing regularization on the NOV region is essential to obtain a plausible image stitching result, and we experimentally find the proper degree of regularization.

\begin{table}[t]
\small
\begin{center}
\caption{PSNR ($\uparrow$) of the overlapping regions on our proposed dataset (synthetic) and UDIS dataset~\cite{uis} (real).}
\begin{tabular}{c||c|c}
\hline
\multirow{3}{*}{Dataset} & \multicolumn{2}{c}{Method} \\
\cline{2-3}
 & Deep homography    & Deep pixel-wise warping \\ \cline{2-3}
                        & UDIS~\cite{uis} v2 (real) & Ours (RAFT) \\ \hline\hline
Ours    & 19.4275       & \textbf{23.5784}      \\ \hline
UDIS~\cite{uis}          & 23.1821       & \textbf{23.5258}   \\ \hline
\end{tabular}
\label{table:quan}
\vspace{-20pt}
\end{center}
\end{table}

\section{More Results}

\subsection{Quantitative Results}
To clarify the superiority of the proposed image stitching method based on pixel-wise warp estimation instead of using homography-based warp estimation, we evaluate the PSNR between the reference image and the warped target image on the OV region. 
Specifically, we compare our method with the recently proposed unsupervised deep image stitching method~\cite{uis} based on the deep homography estimation model. 

Since forward warping process in the proposed pixel-wise warp estimation method leads to inevitable holes in the OV region, we calculate the PSNR on the OV region, excluding the pixels located on the holes.
Note that the input image resolutions for calculating PSNR on our dataset and UDIS dataset~\cite{uis} are $224\times224$ and $256\times256$, respectively.
As shown in Table.~\ref{table:quan}, the proposed framework outperforms the homography-based warp estimation method~\cite{uis} on both the proposed dataset (synthetic) and UDIS dataset~\cite{uis} (real).
Although our framework is trained on the proposed synthetic dataset, our method also achieves superior performance on the real images, which verifies the benefit of the proposed framework.
With the help of precise pixel-wise warps, our pixel-wise deep image stitching method successfully generates plausible stitching results, especially on challenging scenes with large parallax.

\subsection{Qualitative Results}
\begin{figure*}[t]
    \centering
    \includegraphics[width=0.99\linewidth]{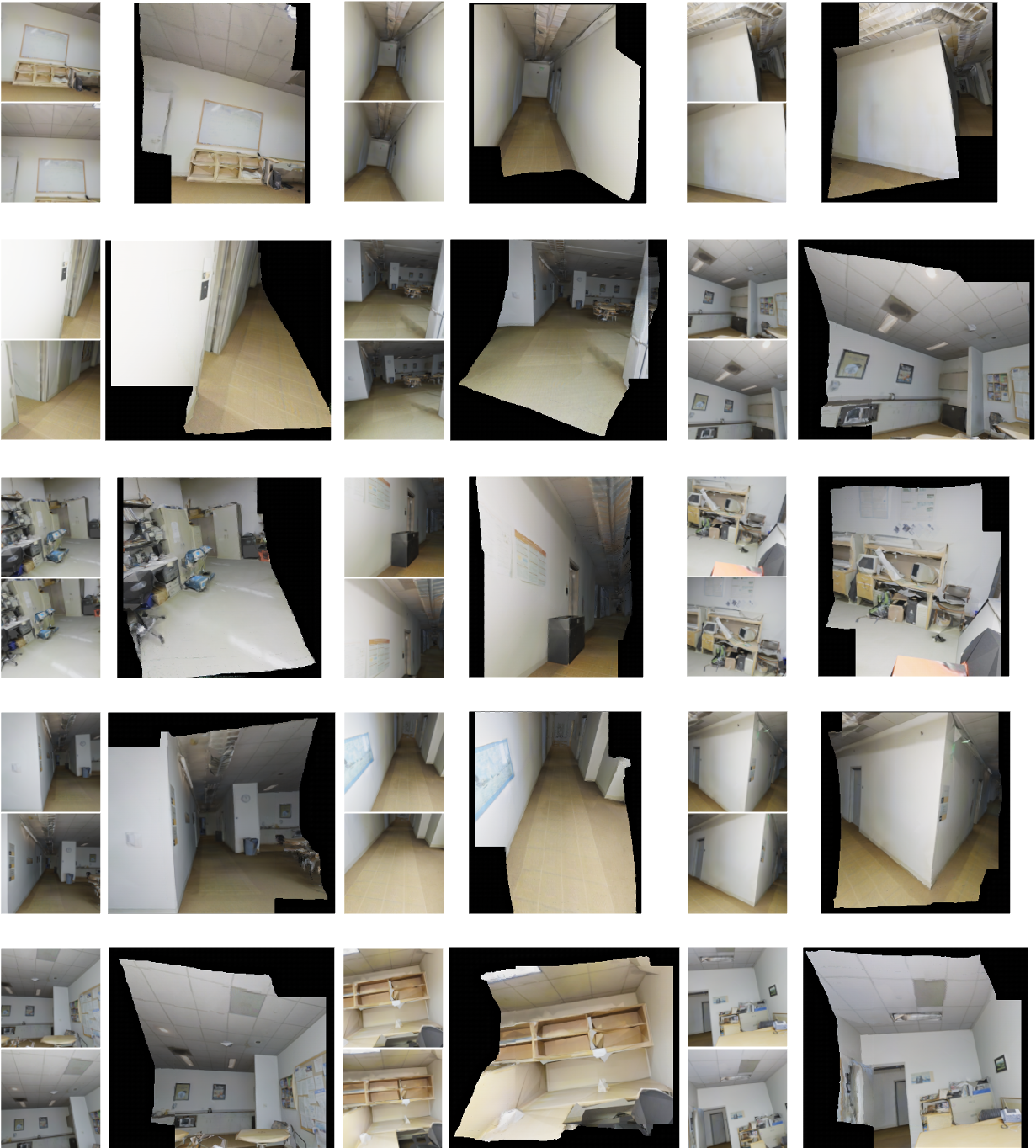}
    \caption{Qualitative results on the proposed dataset.}
    \label{fig:PDIS-supple}
\end{figure*}

\begin{figure*}[t]
    \centering
    \includegraphics[width=0.99\linewidth]{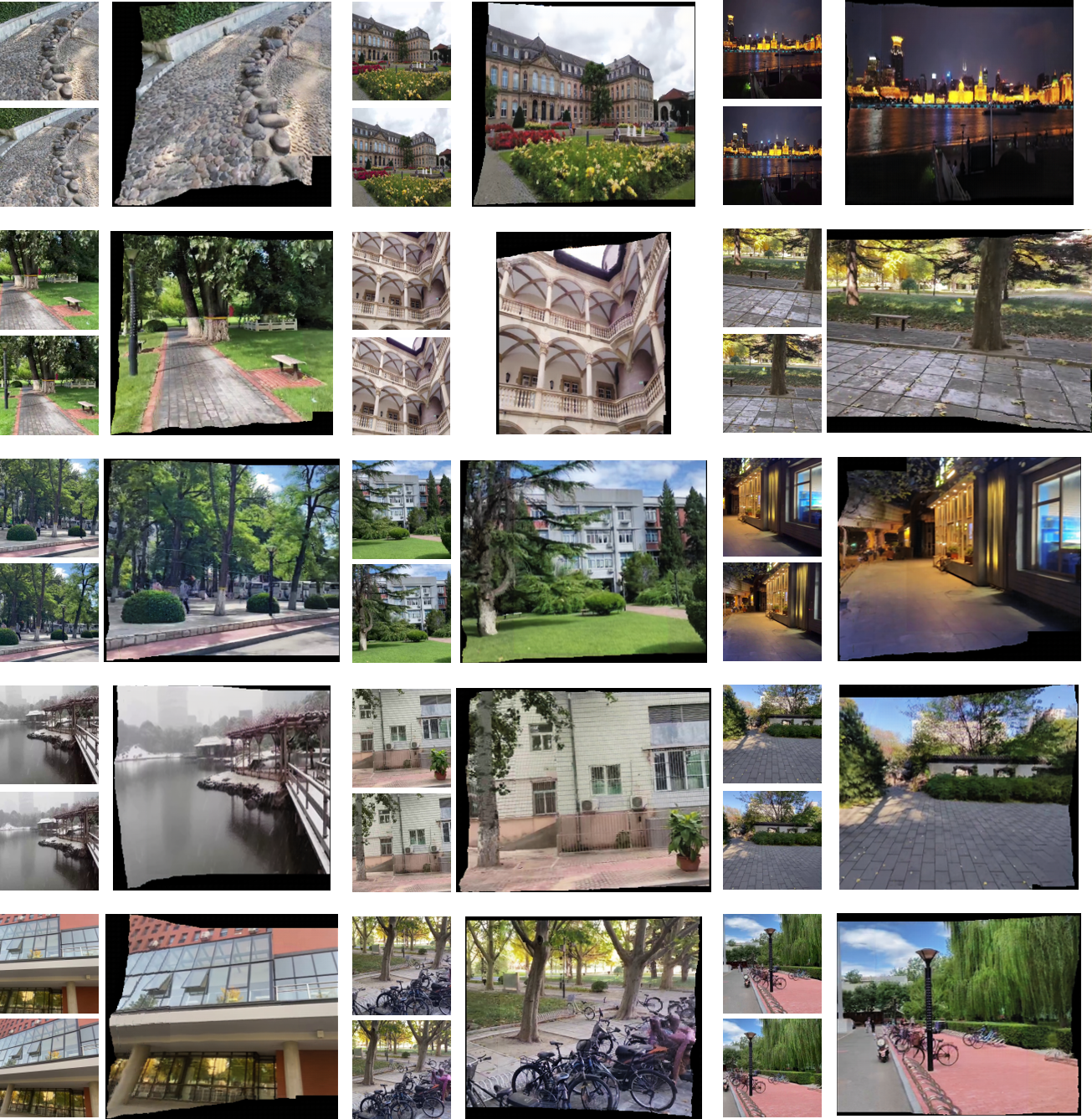}
    \caption{Qualitative results on the UDIS dataset~\cite{uis}.}
    \label{fig:UDIS-supple}
\end{figure*}

\begin{figure*}[t]
    \centering
    \includegraphics[width=0.99\linewidth]{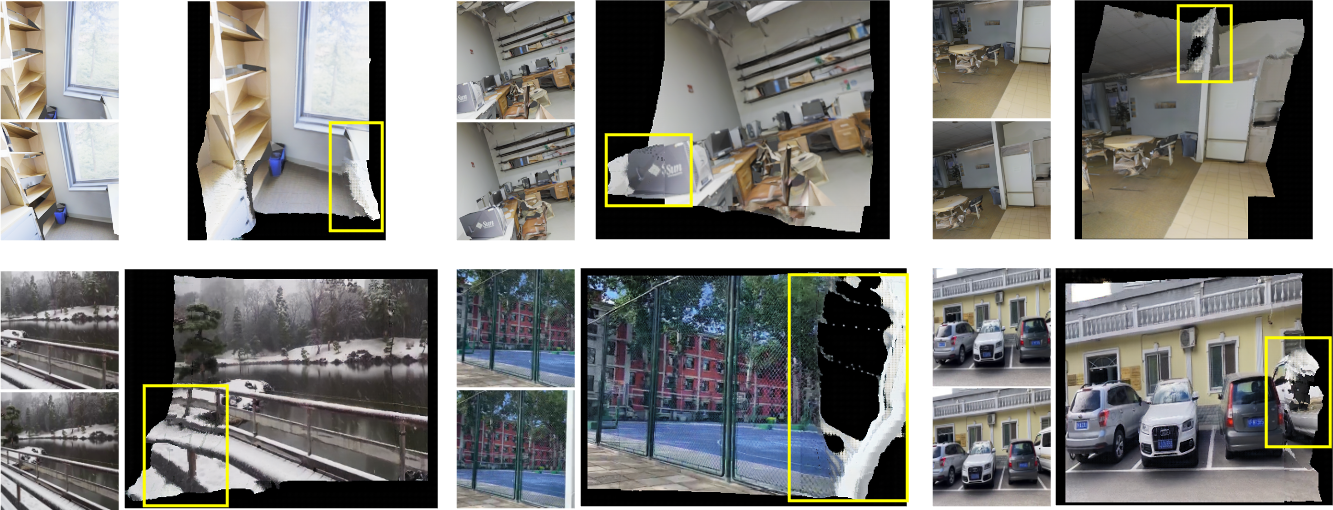}
    \caption{Failure cases on our proposed dataset (1st row) and the UDIS dataset~\cite{uis} (2nd row).}
    \label{fig:failure-supple}
\end{figure*}

We additionally visualize the qualitative results of our model on the proposed dataset and the UDIS dataset~\cite{uis} in Fig.~\ref{fig:PDIS-supple} and Fig.~\ref{fig:UDIS-supple}, respectively.
We also visualize more failure cases of our model on the proposed dataset and the UDIS dataset~\cite{uis} in Fig.~\ref{fig:failure-supple}.

\section{Reproducibility}
We share our PyTorch Lightning implementations and dataset for the reproducibility. However, due to the memory limitation of Microsoft CMT, we could not provide our full dataset. Instead, we provide the \textit{subset of the proposed dataset (10 samples for each area)} and \textit{training} and \textit{inference} codes for PWM and SIGMo. Our method cannot be fully reproduced with the shared materials, but reviewers could check the minimum validity.

\end{appendix}

\end{document}